\def\year{2021}\relax
\documentclass[letterpaper,dvipsnames]{article} % DO NOT CHANGE THIS
\usepackage{aaai21}  % DO NOT CHANGE THIS
\usepackage{times}  % DO NOT CHANGE THIS
\usepackage{helvet} % DO NOT CHANGE THIS
\usepackage{courier}  % DO NOT CHANGE THIS
\usepackage[hyphens]{url}  % DO NOT CHANGE THIS
\usepackage{graphicx} % DO NOT CHANGE THIS
\usepackage{natbib}  % DO NOT CHANGE THIS AND DO NOT ADD ANY OPTIONS TO IT
\usepackage{caption} % DO NOT CHANGE THIS AND DO NOT ADD ANY OPTIONS TO IT
\usepackage{graphicx}
\usepackage{tabularx}
\usepackage{wrapfig}
\usepackage[normalem]{ulem}
\usepackage{amsmath}
\usepackage{subcaption}
\usepackage{xcolor}
\usepackage{multirow}
\usepackage{cancel}
\usepackage{multicol}
\usepackage{makecell}
\usepackage{comment}
\usepackage{amssymb}
\def\squiggly{\bgroup \markoverwith{\textcolor{red}{\lower3.5\p@\hbox{\sixly \char58}}}\ULon}

\newcommand{\U}[1]{\underline{#1}}

\usepackage{url}
\usepackage{xcolor} 
\usepackage{paralist}

\makeatletter
\newcommand{\thickhline}{%
    \noalign {\ifnum 0=`}\fi \hrule height 1pt
    \futurelet \reserved@a \@xhline
}
\makeatother

\title{Unsupervised Opinion Summarization with Content Planning}
\author {
    Reinald Kim Amplayo,
    Stefanos Angelidis,
    Mirella Lapata  \\
}
\affiliations {
	Institute for Language, Cognition and Computation \\
	School of Informatics, University of Edinburgh \\
    reinald.kim@ed.ac.uk, s.angelidis@ed.ac.uk, mlap@inf.ed.ac.uk
}

\begin{document}
\maketitle
\begin{abstract}

  The recent success of deep learning techniques for abstractive
  summarization is predicated on the availability of large-scale
  datasets.  When summarizing reviews (e.g.,~for products or movies),
  such training data is neither available nor can be easily sourced,
  motivating the development of methods which rely on \emph{synthetic}
  datasets for supervised training.  We show that explicitly
  incorporating \emph{content planning} in a summarization model not
  only yields output of higher quality, but also allows the creation
  of synthetic datasets which are more natural, resembling real world
  document-summary pairs.  Our content plans take the form of aspect
  and sentiment distributions which we induce from data without access
  to expensive annotations. Synthetic datasets are created by sampling
  pseudo-reviews from a Dirichlet distribution parametrized by our
  content planner, while our model generates summaries based on input
  reviews and induced content plans.  Experimental results on three
  domains show that our approach outperforms competitive models in
  generating informative, coherent, and fluent summaries that capture
  opinion consensus.

%  Sampling from a DirichletQ distribution reviews, which are
%  automatically induced by a content plan induction model.  These
%  distributions are used as Dirichlet distribution parameters to
%  create a synthetic dataset that closely follows the natural
%  variation of aspect and sentiment mentions in real data.  We then
%  train a summarization model that generates a summary based on both
%  the input reviews and the induced content plan.  We conducted
%  experiments on three different domains: Rotten Tomatoes, Yelp, and
%  Amazon. Our results show that our method outperforms all competing
%  models in generating informative, coherent, and fluent summaries
% capture the opinion consensus.
\end{abstract}

\section{Introduction}
\label{sec:introduction}

\begin{figure}[t]
\fontsize{9}{9}\selectfont 
\begin{center}
\begin{tabular}{@{~}p{8.2cm}@{~}} \thickhline
\multicolumn{1}{c}{Input Reviews}\\\thickhline
\begin{minipage}[t]{8.2cm}
\vspace*{-.2cm}
\begin{asparadesc}
\item[1.] \textit{Local dive bar experience! Authentic phoenix experience squished behind the starbucks.} Pros: \textcolor{Red}{Decent prices, \$2 mystery shots,} \textit{clean bathroom} ...

\item[2.] \textcolor{Red}{Cheap drinks,} \textcolor{Blue}{awesome bar staff,} \textit{stiff pours} ...

\item[3.] \textcolor{Red}{Cheap drinks, great happy hour (that's ridiculously long and cheap)} ... \textcolor{Blue}{I've only found great bartenders and patrons at this little bar} ...

\item[4.] It's a local bar with \textit{no frills except pool table, bar,} and \textcolor{Green}{friendly people} ... \textit{The sliding glass door with the little beach is what makes this place awesome!!!} ...

\item[5.] \textcolor{Blue}{Bartender was friendly and made great shots,} \textit{but the place was full of regulars who made it impossible to have fun} ...

\item[6.] \textit{Their Christmas decorations rival that of coach house but without the Scottsdale crowd.} \textcolor{Green}{You can find every type of person hanging out here.} \textcolor{Blue}{The staff is friendly} ...

\item[7.] ... \textcolor{Green}{reminds me of back home in the Mid West. Good times and great spot to mingle and meet new people!}

\item[8.] \textcolor{Blue}{Lynn is the reason I continue to come back!! She is personable, fun, and dedicated.}
  \vspace*{-.15cm}
\end{asparadesc}
\end{minipage}\\
\multicolumn{1}{c}{} \\\thickhline
\multicolumn{1}{c}{Opinion Summary} \\ \thickhline
\textcolor{Red}{The drinks here are well priced, especially during happy hour.} \textcolor{Green}{There is a large variety of regulars from various backgrounds and ages. Great place to meet new people.} \textcolor{Blue}{The staff are great they provide a nice judgement free environment and they aren't stingy on the pours.}
\\\thickhline
\end{tabular}
\end{center}
%\end{footnotesize} 
\vspace*{-2.5ex}
\caption{Yelp reviews about a local bar and corresponding
  summary. Aspect-specific opinions are in color (e.g.,
  \textcolor{Red}{drinks}, \textcolor{Green}{guests},
  \textcolor{Blue}{staff}), while less salient opinions are shown in
  \textit{italics}.}
    \label{fig:intro}
\end{figure}

The large volume of online product reviews has led to the
proliferation of automatic methods for digesting their content in
order to facilitate decision making.  The fields of opinion mining and
sentiment analysis \cite{pang2007opinion} have offered various
solutions, ranging from sentiment classification
\cite{pang2002thumbs}, to aspect extraction
\cite{mukherjee2012aspect}, and aspect-based sentiment analysis
\cite{pontiki2016semeval}. Beyond extracting surface-level information
(e.g.,~sentiment labels from reviews), effective summarization systems
\cite{hu2006opinion} are needed to succinctly convey opinions to users, e.g.,~to
condense multiple reviews for a given product and identify which
weaknesses and features to pay attention to.

Due to the absence of opinion summaries in review websites and the
difficulty of annotating them on a large scale, most previous work has
relied on extractive approaches
\cite{ku2006opinion,paul2010summarizing,carenini2013multi,angelidis2018summarizing},
where parts of the input reviews are copied and arranged onto a
summary. More recent methods
\cite{chu2019meansum,amplayo2020unsupervised,bravzinskas2019unsupervised}
focus on generating abstractive summaries which can be more
informative and less redundant compared to cut-and-paste extracts.
They consider an \emph{unsupervised learning} setting where there are
only documents (product or business reviews) available without
corresponding summaries. An intuitive solution to the lack of training
data is to create synthetic summary-review pairs
\cite{amplayo2020unsupervised,bravzinskas2019unsupervised} by sampling
a review from a corpus of product reviews, and pretending it is a
summary.

%, an example shown in
%Table~\ref{fig:intro}.

Although synthetic datasets enable the use of supervised training and
have been found to produce higher quality summaries than
autoencoder-based methods \cite{chu2019meansum}, they cannot, by
definition, resemble real-world data.
\citet{bravzinskas2019unsupervised} rely on random sampling to select
the pseudo-summary which might have no connection to the input it
purports to summarize.  \citet{amplayo2020unsupervised} create
multiple input reviews by adding noise to the sampled summary. They
generate syntactically noisy versions or extract lexically similar
reviews under the unrealistic assumption that all reviews with
overlapping vocabulary will be semantically similar to the summary. As
shown in Table~\ref{fig:intro}, real-world reviews discuss a variety
of opinions covering different aspects of the entity under
consideration (e.g.,~for a bar it might be the price of the drinks,
the stuff, the atmosphere of the place).  Some of these aspects are
salient, we expect to see them mentioned in the summary and discussed
in most reviews, while others will be less salient and absent from the
summary. There is also variety among reviews: some will focus on
several aspects, others on a single one, and there will be some which
will discuss idiosyncratic details.

In this paper, we propose to incorporate \emph{content planning} in
unsupervised opinion summarization. The generation literature provides
multiple examples of content planning components
\cite{kukich1983design,mckeown1985text} for various domains and tasks
including data-to-text generation
\cite{gehrmann2018end,puduppully2019data}, argument generation
\cite{hua2019sentence}, and summarization
\cite{kan-mckeown-2002-corpus}. Aside from guiding generation towards
more informative text, we argue that content plans can be usefully
employed to reflect a natural variation of sampled reviews in creating
a synthetic dataset.  Our content plans take the form of aspect and
sentiment probability distributions which are induced from data
without access to expensive annotations.
%Our model consists of three steps.  Firstly, we learn probability distributions from disentangled representations of aspect and sentiment using a content plan induction model. 
Using these as parameters to a Dirichlet distribution, we create a
synthetic dataset of review-summary pairs, where the variation of
aspect mentions among reviews can be controlled.  We also propose an
opinion summarization model that uses these distributions as a content
plan to guide the generation of abstractive summaries.
%We only require review ratings as supervision, which are often attached with the review.

Experiments on three datasets
\cite{wang2016neural,chu2019meansum,bravzinskas2019unsupervised}
representing different domains (movies, business, and product reviews)
and summarization requirements (short vs longer summaries) show that
our approach outperforms competitive systems in terms of ROUGE,
achieving state of the art across the board.  Human evaluation further
confirms that the summaries produced by our model capture salient
opinions as well as being coherent and fluent.

%%%%%%%%%%%%%%%%%%%%%%%%%%%%%%%%%%%%%%%%%%%%%%%%%%%%%%%%%%%%%%%%%%%%%%%%%%%%%%%%%%%%
\section{Related Work}
\label{sec:related-work}

Most previous work on unsupervised opinion summarization has focused
on extractive approaches
\cite{ku2006opinion,paul2010summarizing,carenini2013multi,angelidis2018summarizing}
which cluster opinions of the same aspect or sentiment, and identify
text that represents each cluster.  There have been relatively fewer
attempts to create abstractive summaries.  \citet{ganesan2010opinosis}
generate summaries from textual graphs while other work
\cite{carenini2006multi,difabbrizio2014hybrid} employs a two-stage
framework that first selects salient text units and then generates an
abstractive summary based on templates.

The majority of eural summarization models
\cite{rush2015neural,see2017get} make use of the very successful
encoder-decoder architecture \cite{sutskever2014sequence}, often
enhanced with attention \cite{bahdanau2015neural} and copy mechanisms
\cite{vinyals2015pointer} which have been shown to encourage diversity
and fluency in the output. Unsupervised text generation methods
\cite{freitag2018unsupervised,fevry2018unsupervised,chu2019meansum}
conventionally make use of variational autoencoders
\cite{kingma2014auto}, while employing relatively simple decoders in
order to mitigate posterior collapse
\cite{kingma2014auto,bowman2016generating}.  A more recent line of
work \cite{bravzinskas2019unsupervised,amplayo2020unsupervised}
creates synthetic datasets in cases where gold standard summaries are
not available which in turn allow to train models in a supervised
setting and make use of of effective decoding techniques such as
attention and copy.  Our method is in line with this work, but
ultimately different in its use of content planning to guide both
summarization and synthetic data creation.

%is different
%since we create a more natural dataset and includes content planning
%when generating the summary.

%There were also attempts in generating aspect-specific summaries but relied on supervision from human-annotated summaries \cite{amplayo2019informative} and sentence-level aspect and sentiment labels \cite{coavoux2019unsupervised}. Our work differs in two respects: we rely only on review texts and their ratings for unsupervised training, and we provide a dataset for empirical evaluation among systems.

Content plans have been successfully used to improve generation
performance in both traditional
\cite{kukich1983design,mckeown1985text} and neural-based systems
\cite{gehrmann2018end,puduppully2019data}. Content plans are often
discrete and designed with a specific task and domain in
mind. Examples include a sequence of facts for data-to-text generation
\cite{gehrmann2018end,moryossef2019step}, a list of Wikipedia
key-phrases for argument generation \cite{hua2019sentence}, and entity
mentions and their clusters in news summarization
\cite{amplayo2018entity,sharma2019entity}. Our content plans are
neither discrete nor domain-specific. They take the form of aspect and
sentiment distributions, and serve the dual purpose of creating more
naturalistic datasets for model training and guiding the decoder
towards more informative summaries.

%%%%%%%%%%%%%%%%%%%%%%%%%%%%%%%%%%%%%%%%%%%%%%%%%%%%%%%%%%%%%%%%%%%%%%%%%%%%%%%%%%%

\section{Problem Formulation}
\label{sec:problem-formulation}

We assume access to a collection of reviews about a specific entity
(e.g., a movie, product, business).  
These reviews have ratings, 
which suggest the overall \textit{sentiment} of the reviews and
can be either binary (e.g.,~positive or negative) or on a scale
(e.g.,~from 1 to 5). We further assume that reviews typically focus on
certain \textit{aspects} of the entity, which are features
subject to user opinions (e.g.,~the price and image quality of a
television, the acting and plot of a movie). 
Finally, we do not assume
access to gold-standard summaries, since in most domains these do not
exist.

Let $\mathbf{X}=\{\mathbf{x}_i\}$ denote the set of reviews about an
entity.  The goal of opinion summarization is to generate a summary
$\mathbf{y}$ that covers salient opinions mentioned in the majority of
the reviews.  For each review, we first induce aspect and sentiment
probability distributions~$p(a)$ and~$p(s)$. We do this with a content
plan induction model which learns to reconstruct the review from
aspect and sentiment embeddings.  Distributions $p(a)$ and $p(s)$ are
then used to create a synthetic dataset
$\mathbb{D}=\{\mathbf{X}, \mathbf{y}\}$ of review-summary pairs.  We
make use of the Dirichlet distribution parameterized with $p(a)$ and
$p(s)$ for sampling, which ensures that the reviews are naturally
varied and the summary is representative the opinions found in
reviews.  Finally, we generate opinion summary $\mathbf{y}$ using a
summarization model, which is conditioned on the input reviews
$\mathbf{X}$, but also guided by distributions $p(a)$ and $p(s)$,
which we view as a content plan. %\textbf{synthetic data creation not
  %explained}

%%%%%%%%%%%%%%%%%%%%%%%%%%%%%%%%%%%%%%%%%%%%%%%%%%%%%%%%%%%%%%%%%%%%%%%%%%%%%%%%%%%%%%
\subsection{Content Plan Induction}
\label{sec:cont-plan-induct}

\begin{figure}
	\centering
	\includegraphics[width=\columnwidth]{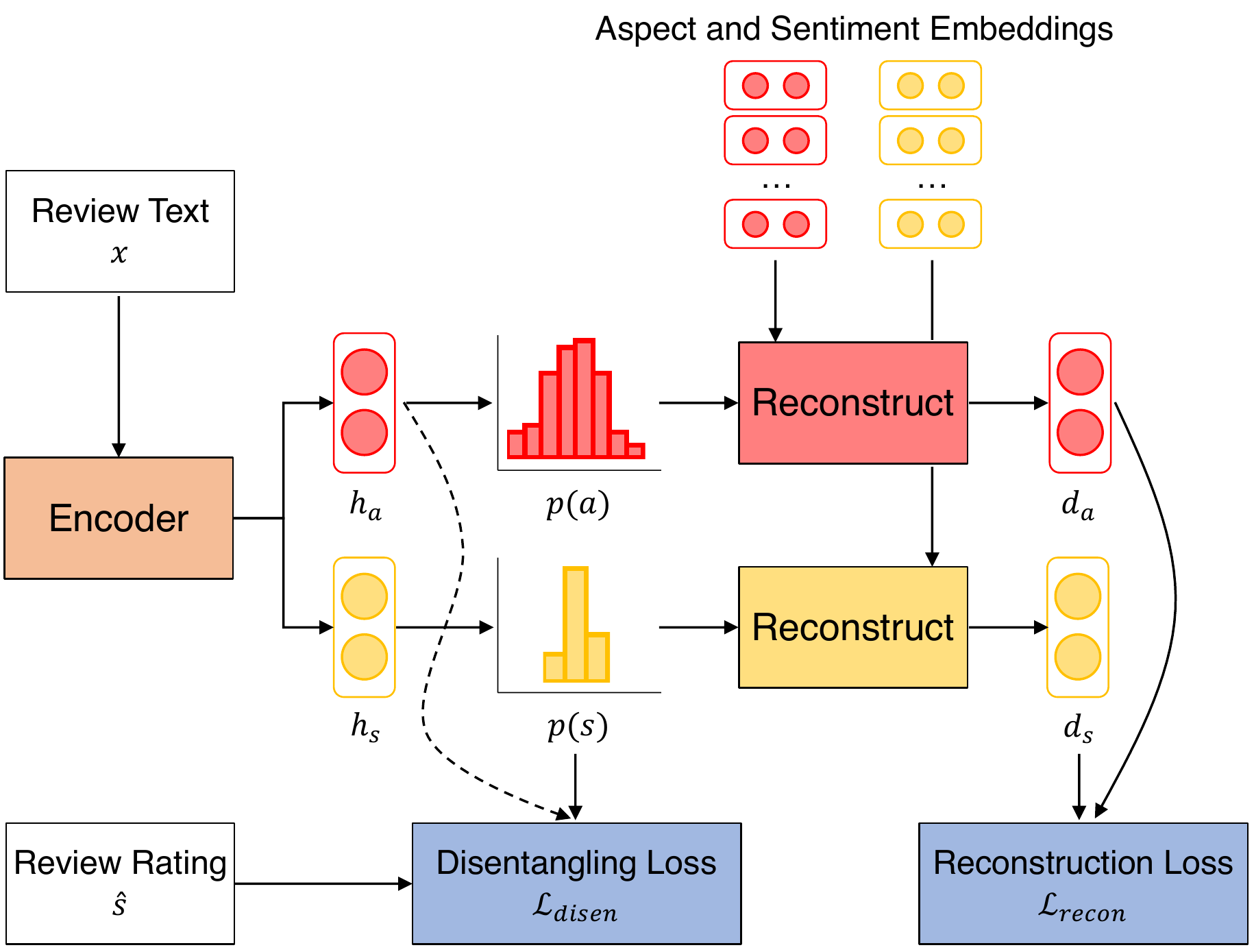}
	\caption{Model architecture of our content plan induction
          model. The dotted line indicates that a reverse gradient
          function is applied.}
	\label{fig:asset}
\end{figure}

Our content plan induction model is illustrated in
Figure~\ref{fig:asset}. It induces probability distributions~$p(a)$
and~$p(s)$ from review $\mathbf{x}$ by learning aspect and sentiment
embeddings, and reconstructing the encoding of $\mathbf{x}$ through
these embeddings. It is similar to neural topic models for aspect
extraction \cite{he2017unsupervised,angelidis2018summarizing}, but
also learns sentiment representations.

We encode review $x=\{w_1,...,w_N\}$ using a neural BiLSTM
\cite{hochreiter1997long} followed by a mean pooling operation. The
output encoding is split into aspect- and sentiment-specific document
encodings, $h_a$ and $h_s$, respectively, which are used in softmax
classifiers to obtain distributions $p(a)$ and $p(s)$ (see
Figure~\ref{fig:asset}):
\begin{align}
	\label{eq:encoder}
	\{h_i\} &= \text{BiLSTM}(\{w_i\}) \\
	h_a, h_s &= \sum\nolimits_i{h_i}/N \\
	p(a) &= \text{softmax}(W_{a} h_a + b_{a}) \\
	p(s) &= \text{softmax}(W_{s} h_s + b_{s})
\end{align}
where~$N$ is the number of review tokens, and W$_a$ and W$_s$
  are weight matrices.

We learn aspect and sentiment embedding
matrices $\mathbf{A}$ and $\mathbf{S}$, via reconstructing the
review. We obtain reconstructions $d_a$ and $d_s$ by weight-summing
embeddings using $p(a)$ and $p(s)$:
\begin{align}
	\label{eq:memory1}
   d_a &= \sum\nolimits_i \mathbf{A}_i * p(a_i) \\
	\label{eq:memory2}
   d_s &= \sum\nolimits_i \mathbf{S}_i * p(s_i)
\end{align}

The model is trained using two different objectives.  Firstly, a
contrastive max-margin objective function is used to reconstruct the
original encodings $h_a$ and $h_s$ with $d_a$ and $d_s$,
respectively. For each review $\mathbf{x}$, we randomly sample $m$
reviews as negative samples and obtain encodings
$\{n^{(i)}_a,n^{(i)}_s\}$ for $1\leq i \leq m$. We formulate the
objective function as a hinge loss $\mathcal{L}_{recon}$ that
maximizes the inner product between $d_a$ and $d_s$ and the original
encodings and minimizes the inner product between $d_a$ and $d_s$ and
the negative samples. We additionally ensure diversity among
aspect/sentiment embeddings in memory \cite{he2017unsupervised} by
adding a regularization term~$\mathcal{R}_{recon}$ to encourage
uniqueness:
\begin{align}
    \mathcal{L}_{recon} &= \sum\nolimits_i \text{max} (0, 1- d_a h_a + d_a n^{(i)}_a) \nonumber \\  
    &+ \sum\nolimits_i \text{max} (0, 1- d_s h_s + d_s n^{(i)}_s) \\
    \mathcal{R}_{recon} &= \|\mathbf{A} \mathbf{A}^\top - \mathbf{I}\| + \|\mathbf{S} \mathbf{S}^\top - \mathbf{I}\|
\end{align}
where $\mathbf{I}$ is the identity matrix. $\mathcal{R}_{recon}$
minimizes the dot product between two different embeddings in memory,
encouraging orthogonality.

We also ensure that the aspect embedding matrix $\mathbf{A}$ does not
include information regarding sentiment, and vice versa, by adding a
disentanglement loss $\mathcal{L}_{disen}$.  This is important since
we want to use aspect information to plan the summary content without
bias towards a certain sentiment.  To distinguish sentiment
information, we leverage review ratings~$\hat{s}$ as sentiment labels
and employ a cross-entropy loss with respect to sentiment
distribution~$p(s)$.  We also predict the same review
ratings~$\hat{s}$ given aspect-specific document encoding~$h_a$ as
input. For this, we use an adversarial classifier with a reverse
gradient function \cite{ganin2016domain} which reverses the sign of
the gradient during backpropagation.
This objective learns the opposite of classifying and thus removes sentiment information
from aspect embeddings~$\mathbf{A}$. We use the following
(adversarial) cross-entropy objective as our disentanglement loss:
\begin{align}
    p(s)_{adv} &= \text{softmax}(\text{GradRev}(W_{adv} h_a + b_{adv})) \nonumber \\
    \mathcal{L}_{disen} &= -\log p(\hat{s}) - \log p(\hat{s})_{adv}
\end{align}

The overall training loss is the linear addition of the reconstruction
and disentanglement losses, and the regularization term mentioned above
($\lambda$ is a hyperparameter controlling the regularization):
\begin{equation}
   \mathcal{L}_{induce} = \mathcal{L}_{recon} + \mathcal{L}_{disen} + \lambda \mathcal{R}_{recon}
\end{equation}

After training, we obtain probability distributions $p(a)$ and $p(s)$
for each review, and use them to create a synthetic dataset and train a summarization model.

%%%%%%%%%%%%%%%%%%%%%%%%%%%%%%%%%%%%%%%%%%%%%%%%%%%%%%%%%%%
\subsection{Synthetic Dataset Creation}
\label{sec:create_data}

To create synthetic dataset $\mathbb{D}=\{\mathbf{X},\mathbf{y}\}$, we
first sample a review from the corpus and pretend it is summary
$\mathbf{y}$. Next, we sample a set of reviews~$\mathbf{X}$
conditioned on $\mathbf{y}$ and pretend they serve as the input which
led to summary~$\mathbf{y}$. We impose a few (stylistic) constraints
on the selection of candidate summaries to ensure that they resemble
actual summaries. We discuss these in our experimental setup.

%the input~$\mathbf{X}$.  Sampling candidate summaries from a corpus of
%reviews requires careful selection since not all reviews are
%summary-worthy.  We follow \citet{amplayo2020unsupervised} by imposing
%a few domain-specific constraints, such as restrictions to length and
%number of non-alphanumeric symbols (listed in Section
%\ref{sec:setup}), to filter out reviews when sampling a summary.

Review samples are created such that they follow the variation of
aspect and sentiment mentions in the sampled summary.  Specifically,
we use a Dirichlet distribution, the conjugate prior of the
multinomial distribution, to sample $N$ pairs of aspect and sentiment
distributions. Given summary~$y$ and its distributions~$p(a)$ and
$p(s)$, the $i$th pair of aspect and sentiment distributions
$\{(p_i(a) p_i(s))\}$, $1 \leq i \leq N$ is sampled as:
\begin{align}
    p_i(a) &\sim \text{Dirichlet}(\alpha_a * p(a)) \\
    p_i(s) &\sim \text{Dirichlet}(\alpha_s * p(s))
\end{align}
where $\alpha_a$ and $\alpha_s$ are constants which control the
variance of the distributions sampled from the Dirichlet. When
$\alpha$ values are small, $p(a)$ and $p(s)$ will look more different
from the distribution of the summary, and when $\alpha$ values are
larger, the sampled distributions will look more similar to the
summary. We provide samples with varying $\alpha$ values in the
Appendix.  Sampling from the Dirichlet ensures that the average of the
sampled distribution equals that of the summary us allowing to control
how the synthetic dataset is created modulating how aspect and
sentiment are represented.

\begin{figure*}[t]
	\centering
	\includegraphics[width=\textwidth]{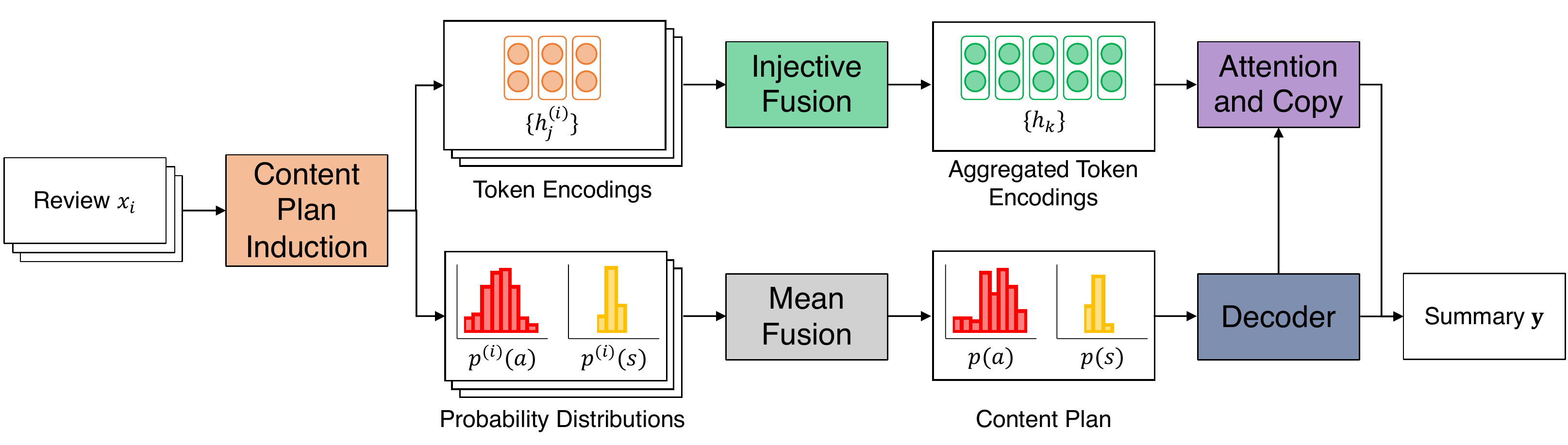}
  \vspace*{-0.7cm}
	\caption{Model architecture of
          \textsc{PlanSum}. The content plan is constructed as the average of the aspect and sentiment probability distributions induced by the content plan induction model. It is then passed to the decoder, along with the aggregated token encodings to generate the summary.}
	\label{fig:PlanSum}
\end{figure*}

Finally, for each sampled pair $(p_i(a), p_i(s))$, we run a nearest
neighbor search over the corpus to find the review $\mathbf{x}_i$ with
the most similar pair of distributions. We use
\citet{hellinger1909neue} distance to quantify the similarity between
two distributions,
i.e..~$sim(p, q) = \|\sqrt{p} - \sqrt{q}\|_2 / \sqrt{2}$ (we take the
average of the similarity scores between aspect and sentiment
distributions).  This results to an instance within dataset $\mathbb{D}$,
where $\mathbf{X}=\{x_1,...,x_N\}$ is the set of reviews for summary
$\mathbf{y}$. We repeat this process multiple times to obtain a large-scale training
dataset.

\subsection{Opinion Summarization}
\label{sec:summarize}

We use the synthetic dataset $\mathbb{D}$ to train our summarization
model which we call \textsc{PlanSum} and illustrate in
Figure~\ref{fig:PlanSum}. A fusion module aggregates token-level
encodings in input reviews~$\mathbf{X}$ to reduce the number of
tokens. The fused encodings are then passed to a decoder that uses the
mean aspect and sentiment distributions as a content plan to generate
output summary~$\mathbf{y}$. We do not employ an encoder in our model,
but rather reuse the encodings from the content plan induction model,
which improves memory-efficiency in comparison to related
architectures
\cite{chu2019meansum,bravzinskas2019unsupervised,amplayo2020unsupervised}.
At test time, the same model is used to summarize actual reviews.

\paragraph{Mean and Injective Fusion} For each review
$\mathbf{x}_i \in \mathbf{X}$ with tokens $\{w^{(i)}_j\}$, we obtain
token-level encodings $\{h^{(i)}_j\}$ and probability distributions
$p^{(i)}(a)$ and $p^{(i)}(s)$, using Equation~\eqref{eq:encoder}. We
then aggregate these encodings and distributions to collectively
represent the set of input reviews.

It is trivial to aggregate aspect and sentiment distributions since the
synthetic dataset is by construction such that their average equals to
the summary. We thus take their mean as follows:
\begin{align}
	\label{eq:plan1}
	p(a) &= \sum\nolimits_i p^{(i)}(a) / N \\
	\label{eq:plan2}
	p(s) &= \sum\nolimits_i p^{(i)}(s) / N 
\end{align}

It is critical to fuse token embeddings as the number of input tokens
can be prohibitively large causing out-of-memory issues.  We could
fuse token embeddings by aggregating over the same word, especially
since multiple reviews are highly redundant. However, simple
aggregation methods such as mean and max pooling may be all too
effective at eliminating redundancy since they cannot retain
information regarding token frequency. This would be problematic for
our task, redundancy is an important feature of opinion summarization,
and repetition can indicate which aspects are considered important. To
mitigate this, we borrow a fusion method from graph neural networks
\cite{xu2018how} that uses an injective function, to effectively
discriminate representations of the same token but with different
levels of redundancy:
\begin{equation}
	h_k = \text{MLP} ( e_k + \sum\nolimits_{(i,j):w^{(i)}_j=w_k} h^{(i)}_j )
\end{equation}
where $e_k$ is a learned embedding for word $w_k$ in the vocabulary.

\paragraph{Decoder with Content Planning} Our decoder is an LSTM
equipped with attention \cite{bahdanau2015neural} and copy
\cite{vinyals2015pointer} mechanisms, where the aggregated token
embeddings $\{h_k\}$ are used as keys. Additionally, at each timestep,
the decoder makes use of the aggregated probability distributions
$p(a)$ and $p(s)$ as a content plan. This guides the model towards
generating correct aspect and sentiment information. Specifically, we
use embedding matrices $\mathbf{A}$ and $\mathbf{S}$ from the content
plan induction model to obtain aspect and sentiment encodings~$d_a$
and~$d_s$, using Equations \eqref{eq:memory1} and
\eqref{eq:memory2}. We then combine these encodings with the output
token $y_t$ at timestep $t$:
\begin{align}
	y'_t &= f(d_a, d_s, y_t) \\
	s_t &= \text{LSTM}(y'_t, s_t) \\
	p(y_{t+1}) &= \textsc{AttendCopy}(y'_t,s_t,\{h_k\})
\end{align}
where $f(\cdot)$ is a linear function.

\paragraph{Training and Inference} We use a maximum likelihood loss to
optimize the probability distribution based on summary
$\mathbf{y} = \{y_t\}$. We also use an LM-based label smoothing
method, which instead of the uniform distribution
\cite{szegedy2016rethinking} uses predictions from BERT
\cite{devlin2019bert} as a prior distribution:
\begin{align}
	\hat{y}_t &= (1-\delta) * y_t + \delta * \text{BERT}(y_{-t}) \\
	\mathcal{L}_{gen} &= -\sum\nolimits_t \hat{y}_t \log p(y_t)
\end{align}

\section{Experimental Setup}
\label{sec:setup}

\subsection{Datasets}

We performed experiments on three opinion summarization
benchmarks. These include the Rotten Tomatoes
dataset\footnote{\url{http://www.ccs.neu.edu/home/luwang/data.html}}
(RT; \citealt{wang2016neural}) which contains a large set of reviews
for various movies written by critics. Each set of reviews has a
gold-standard opinion summary written by an editor. However, we do not
use ground truth summaries for training, to simulate our unsupervised
setting.  Our second dataset is
Yelp\footnote{\url{https://github.com/sosuperic/MeanSum}}
\cite{chu2019meansum} which includes a large training corpus of
reviews for businesses without gold-standard summaries, as well as
development and test sets where summaries were generated by Amazon
Mechanical Turk (AMT) crowdworkers.  Finally, the Amazon
dataset\footnote{\url{https://github.com/ixlan/Copycat-abstractive-Amazon-product-summaries}}
\cite{bravzinskas2019unsupervised} contains product reviews for four
Amazon categories: \textit{Electronics}, \textit{Clothing, Shoes and
  Jewelry}, \textit{Home and Kitchen}, and \textit{Health and Personal
  Care}. The development and test partitions come with three
gold-standard reference summaries produced by AMT annotators.
All datasets include review ratings which we used as sentiment labels: Rotten Tomatoes has binary labels, while Yelp
and Amazon have a 1--5 scale.

To create synthetic training data, we sampled candidate summaries
using the following constraints: (1) there must be no non-alphanumeric
symbols aside from punctuation, (2) there must be no first-person
singular pronouns (not used in Yelp/Amazon), and (3) the number of
tokens must be between 50--90 (20--50 for RT).  We also made sure that
sampled reviews and candidate summary discuss the same entity.  After
applying these constraints we obtained 100k (Yelp), 25k (RT), and 90k
(Amazon) review-summary pairs.  Statistics of these datasets are
reported in Table~\ref{tab:datasets}. As can be seen, RT contains the
largest number of input reviews but the shortest summaries (22--35
tokens). While Amazon and Yelp have a smaller number of input reviews 
but longer summaries (66--70.9 and 62.5--59.8 tokens, respectively).

\begin{table}[t]
	\small
    \centering
    \begin{tabular}{@{}lrrr@{}}\thickhline
      Yelp  & Train* & Dev & Test \\
        \thickhline
        \#summary & 100k & 100 & 100 \\
        \#reviews & 8.0 & 8.0 & 8.0 \\
        \#tokens/summary & 66.0 & 70.9 & 67.3 \\
        \#tokens/review & 65.7 & 70.3 & 67.8 \\
        corpus size & \multicolumn{3}{r@{}}{2,320,800} \\
        \thickhline
      Rotten Tomatoes    &    Train* & Dev &  Test \\      \thickhline
        \#summary & 25k  & 536 & 737 \\
        \#reviews &  72.3 & 98.0 & 100.3 \\
        \#tokens/summary & 25.8 & 23.6 & 23.8 \\
        \#tokens/review & 22.9 & 23.5 & 23.6 \\
   corpus size & \multicolumn{3}{r@{}}{245,848} \\
        \thickhline
      Amazon  & Train* & Dev & Test \\
        \thickhline
        \#summary & 90k & 28$\times$3 & 32$\times$3 \\
        \#reviews & 8.0 & 8.0 & 8.0 \\
        \#tokens/summary & 59.8 & 60.5 & 62.5 \\
        \#tokens/review & 55.8 & 56.0 & 56.0 \\
        corpus size & \multicolumn{3}{r@{}}{1,175,191} \\
        \thickhline
    \end{tabular}
  \vspace*{-0.7ex}
    \caption{Dataset statistics; Train* column refers to the synthetic
      data we created. Amazon contains
      three reference summaries ($\times$ 3) per instance.}
    \label{tab:datasets}
\end{table}

\begin{table*}[t]
  \small
  \centering
    \begin{tabular}{@{}lccccccccc@{}}
    \thickhline
    & \multicolumn{3}{c}{Yelp} & \multicolumn{3}{c}{RT} & \multicolumn{3}{c}{Amazon} \\
    \multicolumn{1}{c}{Model} &  R1    & R2    & RL & R1 & R2 & RL & R1 & R2 & RL\\
    \thickhline
   \textsc{LexRank} & 25.50 & 2.64 & 13.37 & 14.88   & 1.94   & 10.50 & 28.74 & 5.47 & 16.75  \\
   \textsc{W2vCent} & 24.61 & 2.85  & 13.81  & 13.93 & 2.10 & 10.81 & 28.73 & 4.97 & 17.45 \\
   \textsc{SnCent} & 25.05  & 3.09  & 14.56  & 15.90 & 2.01 & 11.74 & 30.45 & 5.40 & 17.73 \\
   \textsc{BertCent} & 26.67 & 3.19 & 14.67 & 17.65 & 2.78 & 12.78 & 30.67 & 5.21 & 17.76 \\
    \hline
    \textsc{Opinosis}  & 25.15 & 2.61 & 13.54  & 14.98     &  3.07    & 12.19 & 28.42 & 4.57 & 15.50 \\
    \textsc{MeanSum}  & 28.86  & 3.66  & 15.91  & 15.79 & 1.94 & 12.26  & 29.20 & 4.70 & 18.15  \\
    \textsc{DenoiseSum}  & \U{30.14}  & {4.99}  & {17.65} & \U{21.26} & \U{4.61} & \U{16.27} & --- & --- & ---\\
    \textsc{Copycat}  & 29.47 & \U{5.26} & \U{18.09} & --- & --- & --- & 31.97 & 5.81 & \textbf{20.16} \\
    \textsc{PlanSum} & \hspace*{.15cm}\textbf{34.79}${}^*$ & \hspace*{.15cm}\textbf{7.01}${}^*$ & \hspace*{.15cm}\textbf{19.74}${}^*$ & \hspace*{.15cm}\textbf{21.77}${}^*$ & \hspace*{.15cm}\textbf{6.18} & \hspace*{.15cm}\textbf{16.98}${}^*$ & \hspace*{.15cm}\textbf{32.87}${}^*$ & \hspace*{.15cm}\textbf{6.12}${}^*$ & \U{19.05}\\
    \thickhline
    \end{tabular}%
  \vspace*{-0.7ex}
    \caption{Automatic evaluation on {Yelp}, {RT}, and Amazon
      datasets.  Extractive/Abstractive models shown in first/second
      block. Best systems shown in bold and 2nd best systems are
      underlined; asterisk (*)~means there is a significant difference
      between best and 2nd best systems (based on paired bootstrap
      resampling; $p < 0.05$).}
    \label{tab:generic_results}%
\end{table*}%

\subsection{Training Configuration}
\label{sec:train-conf}
Across models, we set all hidden dimensions to 256, the dropout rate
to 0.1, and batch size to 16. We used the subword tokenizer of BERT
\cite{devlin2019bert}, which has a 30k token vocabulary trained using
WordPiece \cite{wu2016google}. For RT, we follow
\citet{wang2016neural} and add a generic label for movie titles during
training which we replace with the original title during inference.
We used the Adam optimizer \cite{kingma2015adam} with a learning rate
of $3e-4$, $l_2$ constraint of 3, and warmup of 8,000~steps. We also
used dropout \cite{srivastava2014dropout} after every non-linear
function.  For each dataset, we additionally tuned the number of
aspects, regularization parameter $\lambda$, Dirichlet parameters
$\alpha_a$ and $\alpha_s$, label smoothing parameter $\delta$, and
beam search size on the development set.  We performed early stopping
based on the token-level accuracy of the model, again on the
development set. Our model was trained on a single GeForce GTX 1080Ti
GPU and is implemented using PyTorch.\footnote{Our code can be
  downloaded from \url{https://github.com/rktamplayo/PlanSum}.}  A more detailed model
configuration is described in the Appendix.

\subsection{Comparison Systems}

We compared \textsc{PlanSum} to several previously proposed
approaches.  Extractive systems include \textsc{LexRank}
\cite{erkan2004lexrank}, a PageRank-like algorithm that selects the
most salient sentences from the input, and several variants of a
centroid-based \cite{radev2004centroid} baseline which selects as
summary the review closest to the centroid of a group.  Specifically,
we present results with different input representations, such as
in-domain word2vec \cite{mikolov2013distributed} embeddings
(\textsc{W2vCent}; \citealt{rossiello2017centroid}), encodings from
Sentiment Neuron \cite{radford2017learning}, an LSTM-based language
model trained on a large review corpus (\textsc{SnCent};
\citealt{amplayo2020unsupervised}), and encodings from BERT
\cite{devlin2019bert}, a large transformer-based language model
trained using huge amounts of data (\textsc{BertCent}).

Abstractive comparison systems include \textsc{Opinosis}
\cite{ganesan2010opinosis}, a graph-based method that uses token-level
redundancy to generate summaries, \textsc{MeanSum}
\cite{chu2019meansum}, an autoencoder that generates summaries by
reconstructing the mean of review encodings, \textsc{DenoiseSum}
\cite{amplayo2020unsupervised}, a denoising model that treats
non-salient information as noise and removes it to generate a summary,
and \textsc{Copycat} \cite{bravzinskas2019unsupervised}, a
hierarchical variational autoencoder which learns a latent code of the
summary.

\section{Results}

\paragraph{Automatic Evaluation}
We evaluated the quality of opinion summaries using F$_{1}$ ROUGE
\cite{lin-hovy-2003-automatic}.  Unigram and bigram overlap (ROUGE-1
and ROUGE-2) are a proxy for assessing informativeness while the
longest common subsequence (ROUGE-L) measures fluency.

Our results are summarized in Table~\ref{tab:generic_results}. Among
extractive models, \textsc{BertCent} performs best, indicating that
representations from large transformer-based language models can be
used as a simple method to produce good extractive summaries.
Extractive models, however, are consistently worse than neural-based
abstractive models. Amongst the latter, \textsc{PlanSum} performs best
across datasets and metrics save in terms of ROUGE-L on Amazon. The
slight better performance of \textsc{CopyCat} suggests that the use of
a VAE objective may also be beneficial for our model, however we leave
this to future work. Especially on Yelp, we observe a large
improvement, with an increase of 5.32, 1.75, and 1.65 points in
ROUGE-1/2/L over the best comparison systems. 
Our unsupervised model is comparable to the best supervised model
\cite{amplayo2019informative}, performing 0.58 points better on ROUGE-1
and 0.82 points worse on ROUGE-L.
We show examples of
system output for our model and comparison systems in the Appendix.

\begin{table}[t]
	\small
	\centering
	\begin{tabular}{@{}lccc@{}}
	\thickhline
	\multicolumn{1}{c}{Model} & Yelp & RT & Amazon \\
	\thickhline
	\textsc{PlanSum} & 19.74 & 16.98 & 19.05 \\
	\hline
	\quad No disentangling & 18.83 & 16.09 & 18.52 \\
	\quad No regularization & 19.00 & 16.85 & 18.92 \\
	\hline
	\quad Random sampling & 19.22 & 16.61 & 18.70 \\ %
	\quad Similarity sampling & 19.38 & 15.06 & 18.31 \\
	\hline
	\quad No content plan & 19.03 & 16.56 & 18.28 \\
	\quad Mean token fusion & 18.72 & 16.76 & 18.57 \\
	\quad Uniform label prior & 18.80 & 16.77 & 18.94 \\
	\thickhline
	\end{tabular}
          \vspace*{-0.7ex}
        \caption{\textsc{PlanSum} with less expressive plan induction
          (second block), using alternative review sampling methods
          (third block), and without some modules (fourth block). See
          Appendix for more detailed comparisons.}
  \label{tab:ablation}%
\end{table}

  \begin{table}[t]
  \center
\fontsize{9}{9}\selectfont 
\begin{tabular}{@{}p{8.2cm}@{}}\thickhline
 \multicolumn{1}{c}{\textsc{\small PlanSum}}\\\thickhline
      \textcolor{Red}{This is a great place to
  hang out with friends}. \textcolor{Blue}{The staff is very friendly
  and helpful}. \textcolor{Green}{They have a lot of different beers
  to choose from and the beer selection is
  great}. \textcolor{Green}{I'm not a big fan of beers but this place
  has some good selections}. \textcolor{Red}{If you're in the mood for
  a beer and a fun atmosphere, this will be the
  place for you}. \\\thickhline
\multicolumn{1}{c}{\small Random Sampling}\\\thickhline
\textcolor{Red}{This is a great place to hang out with friends and family}. \textcolor{Green}{The beer selection is great}, and \textcolor{Red}{the atmosphere is very nice}. I've been here a few times and have never had a bad experience. \textcolor{Red}{It's a fun place for a group of friends or groups}. \\\thickhline
\multicolumn{1}{c}{\small Similarity Sampling} \\\thickhline
This is \textit{a great place to go if you're in the area}. It's
                 \textcolor{Red}{a cool place for a night out}, but it
                 is well worth it. \textcolor{Red}{The atmosphere is
                 great} and \textcolor{Blue}{the staff is always
                 friendly}. I'm not sure if I will go back. \\ \thickhline
\multicolumn{1}{c}{\small No Plan} \\\thickhline
This is \textcolor{Red}{a great place to hang out with friends}. \textcolor{Blue}{The staff is very friendly} and \textcolor{Green}{the beer selection is great}. \textcolor{Green}{I've had a couple of beers and they have a good selection of beer and beer.} \textit{It's a little pricey but it is worth the wait}. \\\thickhline
\end{tabular}
  \vspace*{-1ex}
  \caption{\label{tab:cont:output} Yelp summaries generated by
    \textsc{PlanSum} and variants thereof. Aspects also mentioned in
    the gold summary (not shown to save space) are  in color
    (\textcolor{Red}{atmosphere}, \textcolor{Blue}{staff}, and
    \textcolor{Green}{beer}), all other aspects are
    \textit{italicized}.}
  \end{table}

We present in Table~\ref{tab:ablation} various ablation studies on the
three datasets, which assess the contribution of different model
components. Our experiments confirm that aspect and sentiment
disentanglement and embedding regularization in the content plan
induction module improve performance. 
Moreover, our dataset creation
method is better than random or similarity sampling. This is
especially the case on Rotten Tomatoes, where there is an
1.92 decrease in ROUGE-L. Rotten Tomatoes differs from Amazon and Yelp in that the
input reviews are multiple (in the excess of 50) and thus contains
more variety which our content planning approach manages to capture
and reproduce in generating the synthetic data.  Finally, we show that
the use of the content plan, injective fusion module, and the LM-based
label smoothing all increase generation performance.

In Table~\ref{tab:cont:output} we show how content planning modulates
summary output. We present a summary produced by \textsc{PlanSum} and
variants without a content plan during synthetic data creation (see
Random and Similarity Sampling) and in the summarization model (No
Plan). Summaries without any planning whatsoever either miss out on
salient aspects, or focus on aspects that do not reach consensus
(i.e.,~aspect mentions absent from the summary).
%
%\textbf{I was
  %thinking, is it possible to show an example of a summary with and
  %without the plan, you sould pick it carefully of course, and perhaps
  %words representative of the aspect and sentiment distributions?}

\paragraph{Human Evaluation}
\label{sec:human-evaluation}

We also conducted a judgment elicitation study using the Amazon
Mechanical Turk crowdsourcing platform. We assessed the quality of
system summaries using Best-Worst Scaling
\cite{louviere2015best}. Specifically, we asked participants to select
the \textit{best} and \textit{worst} among system summaries taking
into account how much they deviated from given input reviews in terms
of four criteria.  The first two criteria assess informativeness and
ask crowdworkers to select a summary based on whether it mentions the
majority of \emph{aspects} discussed in the original reviews and
agrees with their overall \emph{sentiment}.  We also evaluate
summaries in terms of \emph{coherence} (i.e., is the summary easy to
read and does it follow a natural ordering of facts?), and
\emph{grammaticality} (i.e., is the summary fluent?). We randomly selected 30 instances from the test
set. For Rotten Tomatoes, we filtered out instances where the number
of input reviews exceeded~30 so that participants could read the
reviews in a timely fashion.  We collected three judgments for each
comparison. The order of summaries was randomized per participant. A
rating per system was computed as the percentage of times it was
chosen as best minus the percentage of times it was selected as worst.

We compared summaries produced by the \textsc{BertCent} extractive
baseline, our model \textsc{PlanSum}, and two competitive unsupervised
abstractive systems, \textsc{DenoiseSum}
\cite{amplayo2020unsupervised} and \textsc{Copycat}
\cite{bravzinskas2019unsupervised}. We also included human-authored
summaries as an upper bound.
The ratings are reported in Table \ref{tab:human}. Overall, the gold summaries were consistently rated the highest on all criteria. Among the system summaries, \textsc{PlanSum} was rated the best in terms of all criteria, except on sentiment-based informativeness for Amazon, where \textsc{BertCent} was given the highest rating.
\textsc{BertCent} surprisingly was rated higher than the other abstractive systems. We inspected the summaries produced by these systems and found that 
\textsc{Copycat} summaries are more positive-oriented and \textsc{DenoiseSum} summaries contain more grammatical errors, as also reflected in the ratings.
We posit that these errors are possibly due to the use of random sampling and noising functions, respectively, when creating the synthetic dataset.
We show examples of generated summaries in the Appendix.

%%%%%%%%%%%%%%%%%%%%%%%%%%%%%%%%%%%%%%%%%%%%%%%%%%%%%%%%%%%%%%%%%%%%%%%%%%%%%%%%

\begin{table}[t]
  \small
  \center
\begin{tabular}{@{}lrrrr@{}} \thickhline
Yelp             & Asp & Sen & Coh & Gam \\\thickhline
\textsc{BertCent} &  $-$9.0 & $-$1.5 & $-$2.9 & $-$7.4\\
\textsc{DenoiseSum} & $-$11.3 & $-$11.1 & $-$6.5 & $-$10.6 \\
\textsc{Copycat}   & $-$5.8 & $-$15.0 & $-$15.8 & $-$10.0\\
\textsc{PlanSum} & \textbf{3.9} & \textbf{6.9} & \textbf{5.7} & \textbf{7.0}\\
  \textsc{Gold} & 22.2 & 20.7 & 19.4 & 20.9 \\\thickhline
Rotten Tomatoes             & Asp & Sen & Coh & Gam \\\thickhline
\textsc{BertCent} &  $-$8.4 & $-$12.2 & $-$6.9 & $-$4.0$^{\cancel{*}}$\\
\textsc{DenoiseSum} & $-$31.1 & $-$6.9$^{\cancel{*}}$ & $-$25.1 & $-$17.3 \\
\textsc{Copycat}   & --- & --- & --- & $-$10.0\\
\textsc{PlanSum} & \textbf{10.7} & \textbf{1.3} & \textbf{2.2} & \textbf{$-$2.2}\\
  \textsc{Gold} & 28.9 & 20.4 & 29.8 & 23.6 \\\thickhline
Amazon             & Asp & Sen & Coh & Gam \\ \thickhline
\textsc{BertCent} &  $-$10.7 & \textbf{$-$3.1$^{\cancel{*}}$} & $-$7.1 & $-$9.1$^{\cancel{*}}$\\
\textsc{DenoiseSum} & --- & --- & --- & --- \\
\textsc{Copycat}   & $-$9.8 & $-$18.9 & $-$10.2 & $-$12.22\\
\textsc{PlanSum} & \textbf{0.0} & $-$6.4 & \textbf{7.1} & \textbf{$-$1.8}\\
  \textsc{Gold} & 20.4 & 28.4 & 10.2 & 23.1 \\\thickhline
\end{tabular}
\caption{\label{tab:human} Best-worst scaling: aspect- and
  sentiment-based informativeness (Asp and Sen), coherence (Coh),
  grammaticality (Gram). All pairwise differences between
  \textsc{PlanSum} and other systems are significant, except when
  there is an asterisk (\cancel{*}), using a one-way ANOVA with
  posthoc Tukey HSD tests ($p<0.05$).}
\end{table}

\section{Conclusions}
\label{sec:conclusions}

In this work we considered the use of aspect and sentiment
distributions as a content plan for unsupervised opinion summarization
which we argued leads to higher quality summaries and allows for the
creation of naturalistic synthetic datasets. Extensive automatic and
human-based evaluation showed that our model outperforms competitive
systems on three benchmarks with varying characteristics.  In the
future, we plan to explore personalization in opinion summarization,
where the content plan can be used to control generation towards more
aspect- or sentiment-specific information.  We also plan to apply the 
techniques in this paper to domains where documents are longer (e.g., 
news articles).

%We consider the use of aspect and sentiment distribution as content
%plan for unsupervised opinion summarization. A content plan induction
%predicts these distributions and uses them to create a synthetic
%dataset with reviews following a natural variation of aspect/sentiment
%information. Our summarization model, \textsc{PlanSum}, uses the
%distributions as a content plan to guide the generation. At inference
%time, the model is able to generate aspect-specific summaries using
%only aspect-related keywords. Overall, our model outperforms
%competitive systems by a wide margin.  In the future, we plan to
%explore the use of templates for cases where summaries and documents
%are not similarly written.

\section*{Acknowledgments}

We thank the anonymous reviewers for their feedback.  We gratefully
acknowledge the support of support of the European Research Council
(Lapata, award number 681760) The first author is supported by a
Google PhD Fellowship.

\bibliography{aaai21}
\bibliographystyle{aaai21}

\clearpage
\appendix

\section{Appendix}

\subsection{Training Configurations for Reproducibility}

Our model is implemented in Python 3, and mainly uses the following
dependencies: \texttt{torch}\footnote{\url{https://pytorch.org/}} as
the machine learning library,
\texttt{nltk}\footnote{\url{https://www.nltk.org/}} for text
preprocessing,
\texttt{transformers}\footnote{\url{https://huggingface.co/transformers/index.html}}
for their BERT implementation, and
\texttt{py-rouge}\footnote{\url{https://pypi.org/project/py-rouge/}}
for our evaluation. During our experiments, we used machines with a
single GeForce GTX 1080Ti GPU, 4 CPUs and 16GB of RAMs. The average
training time is 20 hours for Yelp, 12 hours for Rotten Tomatoes, and
17 hours for Amazon. In total and excluding the embedding matrices,
there are 423k parameters in the content plan induction model (409k
for Rotten Tomatoes), and 24.6m parameters in \textsc{PlanSum}.  Table
\ref{tab:hyperparam} shows the hyperparameter values that were tuned
based on the token-level accuracy of the model on the development
sets.

\begin{table}[h]
    \centering
    \begin{tabular}{@{}lccc@{}}
        \thickhline
        & Yelp & RT & Amazon \\
        \hline
        number of aspects & 100 & 50 & 100 \\
        regularization constant $\lambda$ & 1.0 & 1.0 & 1.0 \\
        Dirichlet constant $\alpha$ & 10.0 & 1.0 & 10.0 \\
        label smoothing rate $\delta$ & 0.1 & 0.1 & 0.1 \\
        beam search size & 2 & 2 & 2 \\
        \thickhline
    \end{tabular}
    \caption{Hyperparameters used in \textsc{PlanSum} and
      corresponding token accuracy on the development set for the
      three datasets. We use $\alpha = \alpha_a = \alpha_s$ in our
      experiments.}
    \label{tab:hyperparam}
\end{table}

\subsection{Ablation Studies}

We performed ablation studies on seven different versions of
\textsc{PlanSum}: (a) without disentangling aspect and sentiment
embeddings (i.e., $\mathcal{L}_{disen} = 0$), (b) without uniqueness
regularization of embeddings (i.e., $\lambda = 0$), (c) randomly
sampled reviews when creating a synthetic dataset, as in
\citet{bravzinskas2019unsupervised}, (d) sampled reviews that are
lexically similar (using an IDF-weighted ROUGE-1 F1 score, as in
\citealp{amplayo2020unsupervised}) to the candidate summary, (e)
without a content plan, (f) token aggregation using mean fusion
instead of injective fusion, and (g) use of original label smoothing
method \cite{szegedy2016rethinking} where the prior is set to the
uniform distribution. Table~\ref{tab:ablation_full} shows the
ROUGE-1/2/L F1-scores for the full model and variants thereof. The
final model consistently performs better on all metrics, except on
Rotten Tomatoes where it performs slightly worse than the version that
uses mean fusion to aggregate tokens.

\begin{table*}[t]
  \centering
    \begin{tabular}{@{}lccccccccc@{}}
    \thickhline
     & \multicolumn{3}{c}{Yelp}   & \multicolumn{3}{c}{RT}   & \multicolumn{3}{c}{Amazon} \\
    \multicolumn{1}{c}{Model} & R1 & R2 & RL & R1 & R2 & RL & R1 & R2 & RL \\
    \thickhline
    \textsc{PlanSum} & \textbf{34.79} & \textbf{7.01} & \textbf{19.74} & \textbf{21.77} & {6.18} & \textbf{16.98} & \textbf{32.87} & \textbf{6.12} & \textbf{19.05} \\
    \hline
    \quad No disentangling & 32.25 & 5.74 & 18.83 & 20.90 & 5.50 & 16.09 & 31.51 & 5.51 & 18.52 \\
    \quad No regularization & 32.33 & 5.93 & 19.00 & 21.55 & 6.01 & 16.85 & 31.48 & 5.98 & 18.92 \\
    \hline
    \quad Random sampling & 31.54 & 6.34 & 19.22 & 21.37 & 5.36 & 16.61 & 31.32 & 6.10 & 18.70 \\
    \quad Similarity sampling & 32.80 & 6.42 & 19.38 & 19.47 & 3.85 & 15.06 & 31.54 & 5.98 & 18.31 \\
    \hline
    \quad No content plan & 32.30 & 6.69 & 19.03 & 21.19 & 5.84 & 16.56 & 31.32 & 5.81 & 18.28 \\
    \quad Mean token fusion & 31.22 & 5.44 & 18.72 & 21.42 & \textbf{6.40} & 16.76 & 31.77 & 5.62 & 18.57 \\
    \quad Uniform label prior & 32.85 & 6.10 & 18.80 & 21.57 & 6.21 & 16.77 & 31.00 & 5.54 & 18.94 \\
\thickhline
    \end{tabular}%
    \caption{ROUGE-1/2/L F1 scores of our model and versions thereof with less expressive plan induction (second block), using other review sampling methods (third block), and without some modules in the summarization model (fourth block).}
  \label{tab:ablation_full}%
\end{table*}%

\subsection{Dirichlet Constant}

As discussed in our problem formulation, we control the
variance of the distributions sampled from the Dirichlet distribution
using the $\alpha_a$ (for aspect) and $\alpha_s$ (for sentiment)
constants. This means that when $\alpha$~values are smaller, the
sampled distributions will look more different from the distribution
of the summary. Consequently, when $\alpha$ values are larger, the
sampled distributions will look more similar with the distribution of
the summary. Figure~\ref{fig:dirichlet} shows three examples of 
sampled reviews given a candidate summary and with different $\alpha$
values. We also report the average ROUGE scores between the reviews
and the candidate summary. As can be seen,  ROUGE  increases
as the $\alpha$ value increases, which means that the sampled reviews
get more similar to the summary the larger the value is. Another way
to interpret this is that the review sampling becomes random when the
constant approaches zero, while review sampling uses the similarity
function when it approaches infinity.

\subsection{Example Summaries}

We show example summaries produced by multiple systems, including the
best extractive system \textsc{BertCent}, two neural abstractive
systems
\textsc{DenoiseSum}\footnote{\url{https://github.com/rktamplayo/DenoiseSum}}
\cite{amplayo2020unsupervised} and
\textsc{Copycat}\footnote{\url{https://github.com/ixlan/Copycat-abstractive-opinion-summarizer}}
\cite{bravzinskas2019unsupervised} our model \textsc{PlanSum}, in
Figures \ref{fig:example_yelp1}--\ref{fig:example_yelp2} (for Yelp),
Figure \ref{fig:example_rt} (for Rotten Tomatoes), and Figures
\ref{fig:example_amazon1}--\ref{fig:example_amazon2} (for Amazon).
The extractive model \textsc{BertCent} tends to select reviews with
salient information, however these reviews may contain unnecessary
details, which hurts its performance.  Summaries generated by
\textsc{DenoiseSum} sometimes contain incomprehensible sentences which
may be due to the use of noise function during training, while
summaries generated by \textsc{Copycat} are generally shorter and
positive-oriented which can possibly be a consequence of the use of a
randomly created synthetic dataset.  Overall, \textsc{PlanSum}
produces the best summaries which reflect most salient information
from the input reviews.

To show the effectiveness of the content plan, the figures
additionally show summaries produced by versions of \textsc{PlanSum}
without the use of the content plan: (a--b) \textsc{Random} and
\textsc{Similarity}, a version that instead of the plan, uses random
and similarity sampling, respectively, when creating the synthetic
data, and (c) \textsc{NoPlan}, a version that does not incorporate the
content plan in the summarization model. Without the content plan, the
model produces summaries that either lack information regarding
salient aspects, or include information about aspects that do not
reach consensus (i.e., aspect mentions that are not included in the
\textsc{Gold} summary).

\begin{figure*}
    \small
    \centering
    \begin{tabular}{@{}lr@{}}
        \thickhline
        \multicolumn{2}{@{}c@{}}{Candidate Summary} \\
        \multicolumn{2}{@{}p{16cm}@{}}{I bought one for my mother some years ago due to her arthritis. It worked for her. I bought one for myself and then for all my family members. I don't wish to spend much of my life peeling veggies, but when I do it has to be this product. The soft grip is very helpful in avoiding discomfort, fatigue, and pain.} \\
        \thickhline
        \textbf{Reviews when $\alpha = 1.0$} & {Review-Summary ROUGE}: \textbf{24.75/3.06/15.54} \\
        \hline
        \multicolumn{2}{@{}p{16cm}@{}}{1. This is without a doubt the best peeler I've ever used. My hands never got tired, no matter how many apples, potatoes, carrots, or anything else I peeled. I first bought this peeler when I was suffering from carpal tunnel syndrome (about 5 years ago) and I've been using it ever since and the blade is still sharp!} \\
        \multicolumn{2}{@{}p{16cm}@{}}{2. I'm a sucker for shiny, expensive things, so of course I bought the \$27 stainless steel rosle cross peeler. The oxo good grips peeler does a better job than the more expensive rosle at quickly peeling potatoes, squashes, and other tough veggies. It's a workhorse in the kitchen and cleans up easy in the dishwasher. Highly recommended.} \\
        \multicolumn{2}{@{}p{16cm}@{}}{3. I am a sharp and easy to use peeler. I boast a sleek design and wide handle. I glide over carrots, potatoes, and apples with ease. I romance them until I see their bare flesh, and then let you decide what to do with them. I will replace your old, worn out, rusted peeler in an instant to become your go to kitchen tool. Invest in med and enjoy.} \\
        \multicolumn{2}{@{}p{16cm}@{}}{4. This is our second one we have purchased. The first one lasted for many years before it finally got dull. Oxo makes very high quality products. We also purchased one of these for my parents a few years ago and they still comment about how good it is. I would highly recommend this peeler.} \\
        \multicolumn{2}{@{}p{16cm}@{}}{5. I should have gotten one of these years ago. Fits nicely in my hand and peels great. I need to purchase a couple of more for my children.} \\
        \multicolumn{2}{@{}p{16cm}@{}}{6. Had to find replacement for my mother's peeler that I'd used for 20 years. This one did the trick. Nice comfortable handle, sharp and effective. Easy to clean.} \\
        \multicolumn{2}{@{}p{16cm}@{}}{7. I really like this peeler. It is really smooth, easy to clean, and hold fairly well. I do miss the eye removal tool from the standard peelers but the one it provides works well enough. If the eye removal tool was better I would give it five stars} \\
        \multicolumn{2}{@{}p{16cm}@{}}{8. As a vegan, I work with a lot of vegetables. As a result, this peeler is practically an extension of my arm! I love it! I use it regularly and the blade is still sharp. It works perfectly on carrots, potatoes, cucumbers... you name it!} \\
        \thickhline
        \textbf{Reviews when $\alpha = 10.0$} & {Review-Summary ROUGE}: \textbf{26.34/3.46/16.81} \\
        \hline
        \multicolumn{2}{@{}p{16cm}@{}}{1. I am a sharp and easy to use peeler. I boast a sleek design and wide handle. I glide over carrots, potatoes, and apples with ease. I romance them until I see their bare flesh, and then let you decide what to do with them. I will replace your old, worn out, rusted peeler in an instant to become your go to kitchen tool. Invest in med and enjoy.} \\
        \multicolumn{2}{@{}p{16cm}@{}}{2. This is our second one we have purchased. The first one lasted for many years before it finally got dull. Oxo makes very high quality products. We also purchased one of these for my parents a few years ago and they still comment about how good it is. I would highly recommend this peeler.} \\
        \multicolumn{2}{@{}p{16cm}@{}}{3. moved into a new apartment and these are obviously a must have for any cook. Good quality. Wash easily and are just great. Highly recommend and they won't break your wallet!} \\
        \multicolumn{2}{@{}p{16cm}@{}}{4. I got mine years ago and just bought 2 for family members. Super easy to hold, good grip, comfy and sharp blade. Highly recommend!} \\
        \multicolumn{2}{@{}p{16cm}@{}}{5. Bought this from my grandma in the caribbean so she doesn't have to use a kitchen knife to peel her fruits and veggies and waste good meat from her fruits. She love it. It no. Slip grip is great and it peels smoothly..} \\
        \multicolumn{2}{@{}p{16cm}@{}}{6. This swivel peeler works so well!! I even peel mango using this peeler, and it doesn't break the flesh of the fruit. It glides on the fruit or veggie super smooth and the non slip grip handle is great. It cost like \$9 but so worth it because it works so well.} \\
        \multicolumn{2}{@{}p{16cm}@{}}{7. As the man in the house who cooks, I always appreciate good tools. This peeler works precisely like I hoped it would. Glides through even the roughest peels with ease. Liked it so much, I bought one for my mom to replace her archaic one. Go ahead and order one, you won't regret it!} \\
        \multicolumn{2}{@{}p{16cm}@{}}{8. I have been using this peeler for quite some time now. it does it job perfectly well. I use it to peel potato, carrot and ginger skin. The grip is very good. It has not slipped out of my hand once. I wash this in the dishwasher.} \\
        \thickhline
        \textbf{Reviews when $\alpha = 100.0$} & {Review-Summary ROUGE}: \textbf{30.56/4.15/18.93} \\
        \hline
        \multicolumn{2}{@{}p{16cm}@{}}{1. I am a sharp and easy to use peeler. I boast a sleek design and wide handle. I glide over carrots, potatoes, and apples with ease. I romance them until I see their bare flesh, and then let you decide what to do with them. I will replace your old, worn out, rusted peeler in an instant to become your go to kitchen tool. Invest in med and enjoy.} \\
        \multicolumn{2}{@{}p{16cm}@{}}{2. I should have gotten one of these years ago. fits nicely in my hand and peels great. I need to purchase a couple of more for my children.} \\
        \multicolumn{2}{@{}p{16cm}@{}}{3. Bought this from my grandma in the caribbean so she doesn't have to use a kitchen knife to peel her fruits and veggies and waste good meat from her fruits. She love it. It no. slip grip is great and it peels smoothly.} \\
        \multicolumn{2}{@{}p{16cm}@{}}{4. Had to find replacement for my mother's peeler that I'd used for 20 years. This one did the trick. nice comfortable handle, sharp and effective. Easy to clean.} \\
        \multicolumn{2}{@{}p{16cm}@{}}{5. I have been using this peeler for quite some time now. It does it job perfectly well. I use it to peel potato, carrot and ginger skin. the grip is very good. It has not slipped out of my hand once. I wash this in the dishwasher.} \\
        \multicolumn{2}{@{}p{16cm}@{}}{6. I did order two, but now I only have one. My daughter was at my place helping me in the kitchen and was equally impressed at how well this peeler worked. Of course, I gave her my extra one. I must remember to reorder another one for myself.} \\
        \multicolumn{2}{@{}p{16cm}@{}}{7. My wife's nascent arthritis can make it hard for her to grip small handles, but she loves to cook. The oxo products make her very happy.} \\
        \multicolumn{2}{@{}p{16cm}@{}}{8. I don't know, but from now on, I won't. It is my first peeler, so I can hardly compare, but it works very well, very smooth. I had it for 2 weeks, and peeling veggies is now a fast task in my kitchen.} \\
        \thickhline
    \end{tabular}
    \caption{Examples of sampled reviews given a candidate summary,
      when the Dirichlet constant $\alpha$ is varied (Amazon
      dataset). For simplicity, we use the same value for both
      $\alpha_a$ and $\alpha_s$.}
    \label{fig:dirichlet}
\end{figure*}

\begin{figure*}[t]
    \small
  \centering
    \begin{tabular}{@{}lp{13.6cm}@{}}
      \thickhline
    \multicolumn{2}{c}{Reviews} \\
      \multicolumn{2}{@{}p{16cm}@{}}{1. Birdsong is a gem. A true gem! I was over at noda and wandered back and around to birdsong. \textcolor{Blue}{The staff were very friendly} and \textcolor{Red}{I found the bar a bit like home}. \textit{They have a great outdoor area} and, most importantly, \textcolor{Green}{their beer is quality}. I'm generally not a fan of flavored beers. Ipa por vida! But! \textcolor{Green}{Their jalapeno pale ale!? Hello deliciousness}. Seriously. Give it a try.} \\
      \multicolumn{2}{@{}p{16cm}@{}}{2. \textcolor{Green}{Great beer to try! Fun flavors like jalapeno pale ale}. \textcolor{Blue}{The staff inside is nice and friendly}. \textit{I was able to get a t-shirt with no hassle at all}. \textit{The outdoor seating area is wonderful}. \textit{Birdsong is next door to noda}, so you should definitely check it out!} \\
      \multicolumn{2}{@{}p{16cm}@{}}{3. \textcolor{Green}{Had the extra pale ale and loved it.} In fact I loved everything about this place. \textcolor{Red}{The vibe was ideal for a long night of serious causal drinking}. From \textit{the peanuts on the table} to the \textcolor{Blue}{friendly bartenders}, \textcolor{red}{this place just felt homey as soon as you sat on a stool}. But unlike other dive \textcolor{Green}{this bar has delicious beer} and an \textcolor{Red}{a chill atmosphere that really makes the beer go down quick and easy}. I am looking forward to visiting again!} \\
      \multicolumn{2}{@{}p{16cm}@{}}{4. This is a hiden gem.... Reminds me of Asheville, nc \textcolor{Red}{nice happy laid bk plp} and \textcolor{Green}{great beer. The jalepeno pale ale was amazen}..... It drove my sinses in overload. \textcolor{Green}{The smell and taste wrk great for it}, you have got to try!} \\
      \multicolumn{2}{@{}p{16cm}@{}}{5. \textcolor{Green}{Jalapeno pale ale.... maybe a little crazy.... but so good.} I have been going to birdsong since they first opened. \textcolor{Blue}{I have always enjoyed their free will}. \textcolor{Green}{They have made a couple new brews as of late that I sampled and all are really good}. \textit{I love that they are right across the way from noda brewery} and tend to always go to both of them during my visits. \textit{I love the games and the free peanuts}. \textit{For the quality of the beer, I feel the prices are really good}. Hoping to see some additional brews in the future.} \\
      \multicolumn{2}{@{}p{16cm}@{}}{6. We thought that we would give this place a shot considering the positive reviews. However our visit was all but positive. First of all \textit{I don't believe that the air conditioning was working}. I know that in the past some places do this to encourage the beer consumption. I don't necessarily have a problem with this but if thats the case \textcolor{Green}{at least make sure that the beer is cold}. \textcolor{Blue}{I spoke to someone up at bar and she basically said that because the beer is brewed there its not served cold but just slightly colder than room temperature}. If i wanted a warm pint of bitter I'll go to the U.K. needless to say we won't be back.} \\
      \multicolumn{2}{@{}p{16cm}@{}}{7. \textit{Liked the outdoor space} and \textcolor{Green}{the beer is really good}. \textit{Wish they had more seating}} \\
      \multicolumn{2}{@{}p{16cm}@{}}{8. \textcolor{Red}{Great atmosphere! Was a band on the saturday night we were there that was excellent!} \textcolor{Green}{The beers were wonderful as well}, would be back!} \\
      \thickhline
      {\textsc{\textbf{Gold}}} & 
      \textcolor{Red}{If you're looking for a comfortable and inviting bar this is a great place to go}. \textcolor{Green}{They have a lot of unique beers on tap that you will not find anywhere else}. \textcolor{Blue}{The staff here is extremely friendly}, and after just a couple of minutes \textcolor{Blue}{it feels like you are chatting with an old friend}. The next time you want to head out for some drinks give them a shot! \\
      \hline
      {\textsc{\textbf{BertCent}}} & 
      \textit{Had the extra pale ale and loved it}. In fact I loved everything about this place. \textcolor{Red}{The vibe was ideal for a long night of serious causal drinking}. From \textit{the peanuts on the table} to the \textcolor{Blue}{friendly bartenders}, \textcolor{Red}{this place just felt homey as soon as you sat on a stool}. But unlike other dive \textit{this bar has delicious beer} and an \textcolor{Red}{a chill atmosphere that really makes the beer go down quick and easy}. I am looking forward to visiting again! \\
      {\textsc{\textbf{DenoiseSum}}} & 
      \textcolor{Green}{Great beer} and \textit{most of the outdoor seating is wonderful}. \textcolor{Green}{The beer is really good} and \textcolor{Blue}{the staff was friendly}. I wish found this place to go out of my way to go to. \\
      {\textsc{\textbf{Copycat}}} & 
      This place is awesome! \textcolor{Green}{I had the jalapeno pale ale and it was delicious}. \textit{It was a little pricey but worth the trip}. \textcolor{Blue}{The staff was friendly and helpful as well}. I will be back! \\
      \hline
      {\textsc{\textbf{Random}}} & 
      \textcolor{Red}{This is a great place to hang out with friends and family}. \textcolor{Green}{The beer selection is great}, and \textcolor{Red}{the atmosphere is very nice}. I've been here a few times and have never had a bad experience. \textcolor{Red}{It's a fun place for a group of friends or groups}. \\
      {\textsc{\textbf{Similarity}}} & 
      This is \textit{a great place to go if you're in the area}. It's \textcolor{Red}{a cool place for a night out}, but it is well worth it. \textcolor{Red}{The atmosphere is great} and \textcolor{Blue}{the staff is always friendly}. I'm not sure if I will go back. \\
      {\textsc{\textbf{NoPlan}}} & 
      This is \textcolor{Red}{a great place to hang out with friends}. \textcolor{Blue}{The staff is very friendly} and \textcolor{Green}{the beer selection is great}. \textcolor{Green}{I've had a couple of beers and they have a good selection of beer and beer.} \textit{It's a little pricey but it is worth the wait}. \\
      {\textsc{\textbf{PlanSum}}} & 
      \textcolor{Red}{This is a great place to hang out with friends}. \textcolor{Blue}{The staff is very friendly and helpful}. \textcolor{Green}{They have a lot of different beers to choose from and the beer selection is great}. \textcolor{Green}{I'm not a big fan of beers but this place has some good selections}. \textcolor{Red}{If you're in the mood for a beer and a fun atmosphere, this will be the place for you}. \\
      \thickhline
    \end{tabular}%
    \caption{Examples of opinion summaries generated by multiple
      systems on the \textbf{Yelp} dataset. The first and second
      blocks contain input reviews and the human-generated
      \textsc{Gold} summary. The third block contains summaries
      produced by the best extractive system \textsc{BertCent} and two
      abstractive systems \textsc{DenoiseSum} and
      \textsc{Copycat}. The fourth block contains summaries produced
      by \textsc{PlanSum} and versions thereof without the use of the
      content plan during synthetic data creation (\textsc{Random} and
      \textsc{Similarity}) and in the summarization model
      (\textsc{NoPlan}).  Text snippets that mention aspects also
      mentioned in the \textsc{Gold} summary are color-coded
      (\textcolor{Red}{atmosphere}, \textcolor{Blue}{staff}, and
      \textcolor{Green}{beers}), while all other aspects are
      \textit{italicized}.}
  \label{fig:example_yelp1}%
\end{figure*}%

\begin{figure*}[t]
  \small
  \centering
    \begin{tabular}{@{}lp{13.6cm}@{}}
      \thickhline
    \multicolumn{2}{c}{Reviews} \\
      \multicolumn{2}{@{}p{16cm}@{}}{1. \textit{This is a tattoo spot located on the way south end of the strip in what feels like a nearly abandoned strip mall}. \textit{I walked in without an appointment was able to have an artist work on my tattoo right away}. Note: appointments are best. \textcolor{Green}{The front desk staff were less than friendly. But the artists are great}! \textcolor{Red}{The studio is clean and comfy}. Definitely one of the better places I've been to. I'm very pleased with my tattoo and will be coming back for more work.} \\
      \multicolumn{2}{@{}p{16cm}@{}}{2. \textcolor{Green}{It is rare to find a tattoo shop and good artist as quickly as I did. Collin at west coast was awesome! friendly, great tattoo artist and made my visit quick and easy}. Found the shop on yelp so thought I would leave a review in case someone else wants a great experience. They are the perfect place to checkout while in Las Vegas.} \\
      \multicolumn{2}{@{}p{16cm}@{}}{3. \textit{I got everything I wanted and more with my tattoo.} \textcolor{Red}{The shop was clean and organized.} \textit{It is conveniently located right off the I-15 and Silverado Ranch.} \textcolor{Green}{Russ took his time and made sure every detail was exactly the way I wanted it. He was very kind and personable. If you're looking to get a tattoo done by a nice guy and great artist, go see Russ! He doesn't disappoint}! Thanks again!} \\
      \multicolumn{2}{@{}p{16cm}@{}}{4. My husband and I went in for a lock and key tattoo. \textcolor{Green}{We were incredibly thrilled with the work our artist, Colin, did. He was great and drew our vision of them to create something just perfect for us. We will definitely attempt to have Colin do the next tattoo when we come back to Vegas}.} \\
      \multicolumn{2}{@{}p{16cm}@{}}{5. \textcolor{Green}{I've been worked on by just about every artist there, but Jake is my go-to}. \textit{The work this place is putting out reminds me what I love about tattoos - custom artwork}. \textit{Not to mention, the atmosphere and energy of this place is overwhelmingly... comfortable}. This is a great shop all around, \textcolor{Green}{the artists are extremely talented, smart, funny, super sweet, and they're not bad to look at either!}} \\
      \multicolumn{2}{@{}p{16cm}@{}}{6. A friend referred me to west coast after going I will not get my tattoos anywhere else! \textcolor{Green}{Jake is awesome} and \textcolor{Red}{it's the cleanest shop I've seen}. I would only recommend west coast parlor} \\
      \multicolumn{2}{@{}p{16cm}@{}}{7. I was originally a walk-in with a kinda unusual request to have a micro tattoo on my finger... \textcolor{Green}{I was lucky to find Colin! He got exactly what I was looking for. He was amazing in giving advice about placement, color, etc.... He is an amazing artist.} I look forward to going to see Colin to get more great tattoos.} \\
      \multicolumn{2}{@{}p{16cm}@{}}{8. Amazing! \textit{Such an awesome atmosphere} and \textcolor{Green}{friendly people}. Definitely recommend this establishment for tattoos!! West coast f@ckin rocks!} \\
      \thickhline
      {\textsc{\textbf{Gold}}} & 
      This is an amazing tattoo place! \textcolor{Red}{the shop is extremely clean} and \textcolor{Green}{the tattoo artists are very talented}. I don't want to get my tattoos done anywhere else! If you really want a good job done, \textcolor{Green}{I recommend Colin or Jake, they are the best}! \\
      \hline
      {\textsc{\textbf{BertCent}}} & 
      \textit{This is a tattoo spot located on the way south end of the strip in what feels like a nearly abandoned strip mall}. \textit{I walked in without an appointment was able to have an artist work on my tattoo right away}. Note: appointments are best. \textcolor{Green}{The front desk staff were less than friendly. But the artists are great}! \textcolor{Red}{The studio is clean and comfy}. Definitely one of the better places I've been to. I'm very pleased with my tattoo and will be coming back for more work. \\
      {\textsc{\textbf{DenoiseSum}}} & 
      My husband and I went to find a shop and I was looking for a place located in Las Vegas. \textit{The atmosphere was great and friendly}. \textcolor{Red}{The shop was clean}, and \textcolor{Green}{the staff is extremely kind}. I definitely recommend this establishment to anyone. I would only recommend this place to anyone looking for what you're looking for. I will definitely be coming back to this place. \\
      {\textsc{\textbf{Copycat}}} & 
      I've been going to west coast for over a year now and I'm glad to have found west coast tattoo shop. \textcolor{Red}{Everyone is very friendly and professional}. \\
      \hline
      {\textsc{\textbf{Random}}} & 
      This place is amazing! \textcolor{Green}{The artists are very talented} and \textit{the tattoo is very nice}. I've been coming here for years and it's always a great experience! I have been here a few times, and \textcolor{Green}{they are always so friendly and helpful}. \textit{The shop is immaculately located in the middle of the strip, so be prepared to wait for your next tattoo}. \\
      {\textsc{\textbf{Similarity}}} & 
      This is the best place to get a tattoo in Vegas. I've been going to this place for over a year now and \textcolor{Red}{it's always clean} and \textcolor{Green}{the staff is very friendly. the artists are very nice and professional}. If you're looking for a great experience, look no further. \\
      {\textsc{\textbf{NoPlan}}} & 
      This place is amazing! I've been here a few times and have never had a bad experience. \textcolor{Green}{The staff is super friendly} and \textit{the place has a great vibe}. \textcolor{Green}{I love the fact that they have a lot of artists and artists. They also have great customer service and a very friendly staff}. If you are looking for a fun place to get a tattoo, this is the spot to go. \\
      {\textsc{\textbf{PlanSum}}} & 
      This is the best place to get a tattoo in Las Vegas. \textcolor{Green}{I've been here twice and both times I have been to a lot of different artists}. \textcolor{Green}{The staff is very friendly} and \textcolor{Red}{the shop is very clean}. If you are looking for a new shop, I would highly recommend this place. You won't be disappointed. \\
      \thickhline
    \end{tabular}%
    \caption{Examples of opinion summaries generated by multiple
      systems on the \textbf{Yelp} dataset. The first and second
      blocks contain ten input reviews and the human-generated
      \textsc{Gold} summaries. The third block contains summaries
      produced by the best extractive system \textsc{BertCent} and two
      abstractive systems \textsc{DenoiseSum} and
      \textsc{Copycat}. The fourth block contains summaries produced
      by \textsc{PlanSum} and versions thereof without the use of the
      content plan during synthetic data creation (\textsc{Random} and
      \textsc{Similarity}) and in the summarization model
      (\textsc{NoPlan}).  Text snippets that mentioned aspects also
      mentioned in the \textsc{Gold} summary are color-coded
      (\textcolor{Red}{cleanliness} and \textcolor{Green}{staff}),
      while all other aspects are \textit{italicized}.}
  \label{fig:example_yelp2}%
\end{figure*}%

\begin{figure*}[t]
  \small
  \centering
    \begin{tabular}{@{}lp{13.2cm}@{}}
      \thickhline
    \multicolumn{2}{c}{Reviews} \\
      \multicolumn{2}{@{}p{16cm}@{}}{1. A suspense thriller with a sense of pleasurable unease, the film also serves up a juicy slice of human nature.} \\
      \multicolumn{2}{@{}p{16cm}@{}}{2. A small gem of a movie that defies classification and is as thought-provoking as it is funny, scary and sad.} \\
      \multicolumn{2}{@{}p{16cm}@{}}{3. Miller has crafted an intriguing story of maternal instincts and misguided acts of affection.} \\
      \multicolumn{2}{@{}p{16cm}@{}}{4. An engrossing story that combines psychological drama, sociological reflection, and high-octane thriller.} \\
      \multicolumn{2}{@{}p{16cm}@{}}{5. A stylish thriller.} \\
      \multicolumn{2}{@{}p{16cm}@{}}{6. At heart the movie is a deftly wrought suspense yarn whose richer shadings work as coloring rather than substance.} \\
      \multicolumn{2}{@{}p{16cm}@{}}{7. If this movie leaves you cool, it also leaves you intriguingly contemplative.} \\
      \multicolumn{2}{@{}p{16cm}@{}}{8. Works as a decent urban thriller.} \\
      \multicolumn{2}{@{}p{16cm}@{}}{9. Like a Tarantino movie with heart, alias betty is richly detailed, deftly executed and utterly absorbing.} \\
      \multicolumn{2}{@{}p{16cm}@{}}{10. Kiberlain gives an impressive performance that is harshly uncompromising in its presentation of a woman filled with anger, grief and a highly discernible writing talent.} \\
      \thickhline
      {\textsc{\textbf{Gold}}} & 
      Alias Betty works both as a gripping thriller and as a precisely drawn character study. \\
      \hline
      {\textsc{\textbf{BertCent}}} & 
      A small gem of a movie that defies classification and is as thought-provoking as it is funny, scary and sad. \\
      {\textsc{\textbf{DenoiseSum}}} & 
      The visual style and scares cover up as a an original, but it's never less than intriguing. \\
      \hline
      {\textsc{\textbf{Random}}} & 
      The film's episodic ters is a film that hurtss the viewer with the simplicity of the bourgeois and the city. \\
      {\textsc{\textbf{Similarity}}} & 
      It's not a perfect film, but it is a film that raises a lot of ground and redemption. \\
      {\textsc{\textbf{NoPlan}}} & 
      The film's lasting impression is expressed, but it is a movie that'll stay with you afterward. it has a lot of thoughts. \\
      {\textsc{\textbf{PlanSum}}} & 
      The film is a powerfully constructed thriller that is hypnotic, disturbing, unsettling, and darkly funny. \\
      \thickhline \\
      \thickhline
    \multicolumn{2}{c}{Reviews} \\
      \multicolumn{2}{@{}p{16cm}@{}}{1. A charming comedy with enough surprises to counter its lightness.} \\
      \multicolumn{2}{@{}p{16cm}@{}}{2. It is a very positive film in many ways. it argues that just about anybody can be redeemed.} \\
      \multicolumn{2}{@{}p{16cm}@{}}{3. A well-intentioned, warm movie that becomes increasingly saccharine and silly.} \\
      \multicolumn{2}{@{}p{16cm}@{}}{4. It casts a pleasant, amusing and touching spell.} \\
      \multicolumn{2}{@{}p{16cm}@{}}{5. It's all very sweet, but the film goes in too many directions.} \\
      \multicolumn{2}{@{}p{16cm}@{}}{6. A delightful feature that is as charming as its title and as beautiful as its Venetian setting.} \\
      \multicolumn{2}{@{}p{16cm}@{}}{7. A slow-going but very, very sweet movie.} \\
      \multicolumn{2}{@{}p{16cm}@{}}{8. A feel-good movie well-suited for those who don't require roll-in-the-aisle comedies.} \\
      \multicolumn{2}{@{}p{16cm}@{}}{9. This wonderful Italian comedy pays tribute to the deep yearnings we all have for a life of adventure, romance, and intimacy.} \\
      \multicolumn{2}{@{}p{16cm}@{}}{10. Though there's no denying that bread and tulips is just a feel-good movie, it is a delightfully executed, simple, and unassuming film...} \\
      \thickhline
      {\textsc{\textbf{Gold}}} & 
      Bread and Tulips is a sweet-natured comedy offering gentle, escapist entertainment. \\
      \hline
      {\textsc{\textbf{BertCent}}} & 
      A well-intentioned, warm movie that becomes increasingly saccharine and silly. \\
      {\textsc{\textbf{DenoiseSum}}} & 
      It's an incredibly slight tale, of course, and we've seen this movie. \\
      \hline
      {\textsc{\textbf{Random}}} & 
      Bread and Tulips isn't a romantic comedy, but it's also a warm and warm tale that'll be a treat for the ages. \\
      {\textsc{\textbf{Similarity}}} & 
      It's pollyanna, but it is a sweet, delightfully delightful romantic comedy, and a delightful, lonely film. \\
      {\textsc{\textbf{NoPlan}}} & 
      Bread and Tulips is a frostable and hilarious french comedy that survives to be loved by the box office and charm of the workplace. \\
      {\textsc{\textbf{PlanSum}}} & 
      Bread and Tulips is a cute, funny, charming, romantic comedy that is more than a series of fun, and it's a very funny movie. \\
      \thickhline
    \end{tabular}%
    \caption{Examples of opinion summaries generated by multiple
      systems on the \textbf{Rotten Tomatoes} dataset. For each
      instance, the first and second blocks contain input
      reviews and the human-generated \textsc{Gold} summaries. The
      third block contains summaries produced by the best extractive
      system \textsc{BertCent} and the abstractive system
      \textsc{DenoiseSum}. The fourth block contains summaries
      produced by \textsc{PlanSum} and versions thereof without the
      use of the content plan during synthetic data creation
      (\textsc{Random} and \textsc{Similarity}) and in the
      summarization model (\textsc{NoPlan}).}
  \label{fig:example_rt}%
\end{figure*}%

\begin{figure*}[t]
  \small
  \centering
    \begin{tabular}{@{}lp{13.6cm}@{}}
      \thickhline
    \multicolumn{2}{c}{Reviews} \\
      \multicolumn{2}{@{}p{16cm}@{}}{1. \textcolor{Blue}{The only thing I would like to see is an aux cord when I don't want to charge my phone}, but it's not a huge deal. \textcolor{Red}{The sound is great}, and \textit{worth the money}. \textcolor{Blue}{The remote works with your phone, and that's precisely what I wanted}} \\
      \multicolumn{2}{@{}p{16cm}@{}}{2. \textit{While I like the dream machine I don't know why there's so much static}. \textit{It's nearly impossible to get a couple of my favorite radio stations without constant static in the background}. My other radio doesn't do that. I've even tried different locations for it. That's a big disappointment and shortcoming of the product.} \\
      \multicolumn{2}{@{}p{16cm}@{}}{3. \textcolor{Blue}{You need to buy an adaptor for iPod nano's so it was disappointing when my son opened it up on Christmas and could not use it for his iPod nano}. It does not state that anywhere on the box or when i ordered it.} \\
      \multicolumn{2}{@{}p{16cm}@{}}{4. As always, Sony has a 'winner' in this combined am/fm radio and docking station. \textcolor{Red}{Great sound}, \textcolor{Green}{looks good and wife is very pleased as she put it in her craft work area. Finding the combo of am/fm wasn't easy either.} Lots of fm only units. This is a great product.} \\
      \multicolumn{2}{@{}p{16cm}@{}}{5. I was looking quite awhile to locate a decent sounding radio/iPod player which would also charge my iPod. This is perfect for our family. \textcolor{Green}{It's a lot smaller than I thought, which is good.} \textcolor{Blue}{And when we update to an iPhone 5, there is a \$5 adapter to get so we can still use this radio.} Perfect!} \\
      \multicolumn{2}{@{}p{16cm}@{}}{6. \textcolor{Red}{The sound of the radio is of real quality.} \textcolor{Blue}{I also like having the two separate alarms} and \textcolor{Red}{the alarm is not obnoxious yet still wakes us up}. \textcolor{Blue}{My wife charges her iPhone on it regularly and works out well}. We like the sony so much I got one for my son and his wife for a Christmas present} \\
      \multicolumn{2}{@{}p{16cm}@{}}{7. \textit{Love, love, love the ability to save multiple preset radio stations}, and \textcolor{Red}{the sound is clear, crisp...} Amazing! It almost makes waking up a pleasure. \textit{Another feature I never thought I wanted, but really appreciate, is the ability to set the brightness of the clock readout.} Brilliant!} \\
      \multicolumn{2}{@{}p{16cm}@{}}{8. My husband really like this speaker... Love it! \textit{Its so easy to operate by setting the alarm..} \textcolor{Blue}{I like the way when you put your iPhone 4s to the dock its charge at the same time while you are you using it...!} Great product} \\
      \thickhline
      {\textsc{\textbf{Gold}}} & 
      This fm/am radio, iPhone docking station and alarm clock is a perfect combination! \textcolor{Red}{The sound is amazing, the alarm clock is not annoying}, and \textcolor{Green}{the design looks great}! \textcolor{Blue}{It would be nice to have a place to use an aux cord and certain apple products require a \$5 adapter to use the docking station} but other than that, this product is fantastic! \\
      \hline
      {\textsc{\textbf{BertCent}}} & 
      \textcolor{Red}{The sound of the radio is of real quality}. \textcolor{Green}{I also like having the two separate alarms} and \textcolor{Red}{the alarm is not obnoxious yet still wakes us up}. \textcolor{Blue}{My wife charges her iPhone on it regularly and works out well}. We like the Sony so much i got one for my son and his wife for a Christmas present \\
      {\textsc{\textbf{Copycat}}} & 
      Bought this for my daughter for her birthday. She loves it and uses it all the time. \textit{It is easy to set up and use}. I would recommend it to anyone who likes to listen to music. \\
      \hline
      {\textsc{\textbf{Random}}} & 
      This is a great product. \textit{It is very easy to use}, and \textcolor{Red}{the sound quality is great}. \textcolor{Red}{the only complaint i have is that the alarm clock isn't very loud}. i would not recommend this product to anyone. \\
      {\textsc{\textbf{Similarity}}} & 
      I bought this for my wife's iPod nano. \textit{it is very easy to set up}, and \textcolor{Red}{the sound quality is great}. \textcolor{Blue}{The only drawback is that it doesn't have a lot of features to charge it}. I would recommend this product to anyone. \\
      {\textsc{\textbf{NoPlan}}} & 
      This is a great product. \textit{It is easy to use and works great with my iphone 4s}. \textcolor{Green}{the only problem i have is that it's a little bulky}, but i'm not sure if it would have been a problem. i would recommend this player to anyone who is looking for a docking station. \\
      {\textsc{\textbf{PlanSum}}} & 
      \textcolor{Green}{This is a great little radio} \textit{for the price}. \textit{It is easy to use} and \textcolor{Red}{the sound quality is great}. \textcolor{Blue}{The only thing I don't like is that it's not really a dock, since it does not have a cord}. I would recommend this to anyone who wants to listen to music. \\
      \thickhline
    \end{tabular}%
    \caption{Examples of opinion summaries generated by multiple
      systems on the \textbf{Amazon} dataset. The first and second
      blocks contain input reviews and the human-generated
      \textsc{Gold} summaries. The third block contains summaries
      produced by the best extractive system \textsc{BertCent} and the
      abstractive system \textsc{Copycat}. The fourth block contains
      summaries produced by \textsc{PlanSum} and versions thereof
      without the use of the content plan during synthetic data
      creation (\textsc{Random} and \textsc{Similarity}) and in the
      summarization model (\textsc{NoPlan}).  Text snippets that
      mentioned aspects also mentioned in the \textsc{Gold} summary
      are color-coded (\textcolor{Red}{sound quality},
      \textcolor{Green}{design}, and \textcolor{Blue}{accessories}),
      while all other aspects are \textit{italicized}.}
  \label{fig:example_amazon1}%
\end{figure*}%

\begin{figure*}[t]
  \small
  \centering
    \begin{tabular}{@{}lp{13.6cm}@{}}
      \thickhline
    \multicolumn{2}{c}{Reviews} \\
      \multicolumn{2}{@{}p{16cm}@{}}{1. Yes, HP dvd's are dvd's for the better. \textit{Better price}. \textcolor{Green}{Better quality}. \textcolor{Green}{I have used these over the years for many different projects and the quality is there} and \textit{so is the price}. I have had trouble with some other brand named dvd's, but not with HP.} \\
      \multicolumn{2}{@{}p{16cm}@{}}{2. I have had a ton a problems with these discs. \textcolor{Green}{After about 30 minutes of a dvd, it begins to get choppy and become unviewable}. \textit{Looking at the burn side of the disc, there is a area where you can see the burning stopped and i guess picked again}. Do not recommend.} \\
      \multicolumn{2}{@{}p{16cm}@{}}{3. \textcolor{Blue}{Vendor describes the product as being gold in color. It is not it is silver. I know that hp no longer manufactures the gold version. But was hoping this vendor had some gold version of dvd+r in it inventory}. They need change the picture and description to silver instead of gold.} \\
      \multicolumn{2}{@{}p{16cm}@{}}{4. After receiving this dvd+r, \textit{I think I should order the 50 piece spindle}. I did not burn too often. \textit{I believe it is enough for 2 years}. I tried a couple of discs, anyway, like it.} \\
      \multicolumn{2}{@{}p{16cm}@{}}{5. \textcolor{Green}{Ive always bought the HP dvd+r 100 pk and find that they go a long way}, \textcolor{Red}{seldom do i find a coaster in the pack...} hwever his was my first time purchasing the 100 pk from Amazon, but so far so good.} \\
      \multicolumn{2}{@{}p{16cm}@{}}{6. \textcolor{Red}{Produces about 25\% coasters, and another 10-15\% that won't play on dvd players.} \textcolor{Green}{The first half was about what I expected, a few coasters, a couple minor errors. The second half I was lucky to get maybe 5 that worked flawlessly}. No good!} \\
      \multicolumn{2}{@{}p{16cm}@{}}{7. Been purchasing only HP dvd+r media for 8 years or more. Always got the HP branded gold coated. Made by cmc, with id of cmcmagm01001. \textcolor{Blue}{This batch is not gold as portrayed in the photo, it's silver coated, but still the same id}.} \\
      \multicolumn{2}{@{}p{16cm}@{}}{8. \textcolor{Red}{The box was too big and allow the spindle to bounce around in box and plastic covering spindle which keeps the blank dvd in place was broken and dvd were off the spindle. If that is the only size boxes have should put more packing so it will not allow spindle to move at all}.} \\
      \thickhline
      {\textsc{\textbf{Gold}}} & 
      \textcolor{Blue}{These are silver discs, not the gold ones as advertised}. \textcolor{Red}{The packaging is not good, they need to be packed more securely, the dvds shouldn't be able to slide around}. \textcolor{Green}{The quality of the dvds is hit or miss. You might have a good batch and then run into several that won't work}. \\
      \hline
      {\textsc{\textbf{BertCent}}} & 
      After receiving this dvd+r, \textit{I think I should order the 50 piece spindle}. I did not burn too often. \textit{I believe it is enough for 2 years}. I tried a couple of discs, anyway, like it. \\
      {\textsc{\textbf{Copycat}}} & 
      It's a great product. I have had no problems with it and \textit{the price is right}. I would recommend this product to anyone who wants a good quality product. \\
      \hline
      {\textsc{\textbf{Random}}} & 
      \textit{These are great for the price}. \textcolor{Green}{I have used them for about a month now and they have held up very well}. \textit{The only thing I don't like about them is that they are not in the way}, but i'm not sure if they will last. \\
      {\textsc{\textbf{Similarity}}} & 
      I have been using these for a few years now and they are still working great. \textcolor{Green}{I don't know if they will last longer than some of the other brands I've seen}. They don't have a problem with any of them. \\
      {\textsc{\textbf{NoPlan}}} & 
      This is a great product for the money. \textcolor{Green}{It's a good quality product}. \textit{The only thing I don't like about it is that it does not have a lot of space to hold it in place}. i have not had any problems with it, \textcolor{Green}{but i'm sure it will last a long time}. \\ 
      {\textsc{\textbf{PlanSum}}} & 
      I have used these discs for a few years now and have had no problems with them. \textcolor{Green}{They are a bit flimsy, but they don't have a lot of wiggle. I'm not sure if they're going to last longer than a year or so}. \textit{The price is right} and \textcolor{Red}{the packaging is good}. \\
      \thickhline
    \end{tabular}%
\caption{Examples of opinion summaries generated by multiple systems on the \textbf{Amazon} dataset. The first and second blocks contain  input reviews and the human-generated \textsc{Gold} summaries. The third block contains summaries produced by the best extractive system \textsc{BertCent} and the abstractive system \textsc{Copycat}. The fourth block contains summaries produced by \textsc{PlanSum} and versions thereof without the use of the content plan during synthetic data creation (\textsc{Random} and \textsc{Similarity}) and in the summarization model (\textsc{NoPlan}).
  Text snippets that mentioned aspects also mentioned in the \textsc{Gold} summary are color-coded (\textcolor{Red}{packaging}, \textcolor{Green}{quality}, and \textcolor{Blue}{design}), while all other aspects are \textit{italicized}.}
  \label{fig:example_amazon2}%
\end{figure*}%

\end{document}